\documentclass{article}
\usepackage{PRIMEarxiv}
\usepackage[utf8]{inputenc}
\usepackage[T1]{fontenc}
\usepackage[authoryear,round]{natbib}
\setcitestyle{citesep={;},aysep={,},yysep={;}}

\usepackage{amsmath,amsfonts,bm}

\def\eqref#1{equation~\ref{#1}}

\def\1{\bm{1}}

\DeclareMathAlphabet{\mathsfit}{\encodingdefault}{\sfdefault}{m}{sl}
\SetMathAlphabet{\mathsfit}{bold}{\encodingdefault}{\sfdefault}{bx}{n}

\usepackage{hyperref}
\usepackage{url}
\usepackage{graphicx}
\usepackage{booktabs}
\usepackage{multirow}
\usepackage{array}
\usepackage{tabularx}
\usepackage{amsmath}
\usepackage{amssymb}
\usepackage{comment}
\usepackage{float}
\usepackage{mathtools}
\usepackage{bm}
\usepackage{xcolor}
\usepackage{microtype}
\usepackage{enumitem}
\usepackage{pdflscape}

\definecolor{tbdred}{RGB}{170,35,35}

\newcommand{\RR}{\mathbb{R}}

\DeclareMathOperator{\diag}{diag}

\title{MoSPR: Histology-to-Gene Expression Prediction with Morpho-Spatial Macrostates and Low-Rank Molecular Programs}

\author{
  Dongmyung Shin\textsuperscript{\normalfont\ensuremath{*,\dagger}} \\
  OmixAI Co. Ltd. \\
  Oncocross Co. Ltd. \\
  \texttt{shinsae11@omixai.com} \\
  \And
  Geongyu Lee\textsuperscript{\normalfont *} \\
  OmixAI Co. Ltd. \\
  \texttt{gglee@omixai.com} \\
  \AND
  Yesung Cho\textsuperscript{\normalfont *} \\
  OmixAI Co. Ltd. \\
  \texttt{yscho@omixai.com} \\
  \And
  Park Jong Bae\textsuperscript{\normalfont\ensuremath{\dagger}} \\
  Kyunghee University \\
  OmixAI Co. Ltd. \\
  Oncocross Co. Ltd. \\
  \texttt{jbpark@omixai.com} \\
}
\date{}

\hypersetup{hidelinks,pdfauthor={Dongmyung Shin, Geongyu Lee, Yesung Cho, Park Jong Bae},
  pdftitle={MoSPR: Histology-to-Gene Expression Prediction with Morpho-Spatial Macrostates and Low-Rank Molecular Programs}}

\begin{document}
\maketitle
\begingroup
\renewcommand{\thefootnote}{\fnsymbol{footnote}}
\footnotetext[1]{Dongmyung Shin, Geongyu Lee, and Yesung Cho contributed equally.}
\footnotetext[2]{Co-corresponding authors: Dongmyung Shin (\texttt{shinsae11@omixai.com}) and Park Jong Bae (\texttt{jbpark@omixai.com}).}
\endgroup
\setcounter{footnote}{0}
\begin{abstract}
Predicting molecular profiles from histopathology remains challenging because
whole-slide images contain spatially organized, heterogeneous tissue patterns, while gene
expression comprises thousands of correlated targets. We introduce MoSPR
(Morpho-Spatial Program Regression), a linear framework that couples 
an adjacency-informed histology representation with a low-rank molecular basis. 
MoSPR clusters frozen patch embeddings
into morphology microstates, aggregates their spatial adjacencies across the
training cohort, and groups microstates with similar adjacency patterns into
shared macrostates. Each slide is then represented by global morphology and
macrostate-specific deviations, which are linearly mapped to coefficients of 
a training-derived low-rank gene-expression basis. Across three
cancer cohorts from The Cancer Genome Atlas, MoSPR achieves the highest mean
gene-expression prediction scores among all evaluated methods. Without
pathway-level supervision, pathway scores derived from its predicted expression
profiles rank first in eight of nine comparisons across
three pathway collections. Ablation studies on the breast cancer cohort show complementary gains from
adjacency-derived macrostate representation and low-rank molecular prediction. Moreover, with half of the training data on this cohort, MoSPR exceeds the full-data gene-prediction score of the strongest competing baseline. Finally, its linear formulation enables exact decomposition of each predicted expression profile into global and macrostate-specific molecular contributions, providing an
interpretable link between spatially coherent macrostate regions and their
associated molecular programs. Our code is available at \url{https://github.com/Radisen-Panthera/MoSPR}.
\end{abstract}

\section{Introduction}

Molecular profiling is central to cancer diagnosis, biological stratification,
and treatment selection, but it remains substantially more expensive and less
routinely available than hematoxylin-and-eosin (H\&E) histology. This has
motivated methods that infer transcriptomic and other molecular measurements
directly from whole-slide images (WSIs)~\citep{schmauch2020he2rna,
alsaafin2022trnaformer,senbabaoglu2024mosby,pizurica2024sequoia,
nishimura2026cpnn}. A successful model could provide a low-cost virtual
molecular screen, prioritize samples for confirmatory assays, and expose
image--molecule associations~\citep{coudray2018classification, saillard2023msintuit, schmauch2020he2rna}.

Despite rapid progress, histology-to-molecular prediction remains challenging
because WSIs contain heterogeneous tissue patterns, while bulk gene expression
comprises thousands of correlated targets. Global pooling and standard
multiple-instance learning (MIL) summarize overall patch-level morphology but do not
explicitly encode the spatial adjacency of recurring tissue states.
Transformer- and graph-based methods~\citep{shao2021transmil,
chen2021patchgcn} can model contextual or spatial interactions, but operate
over patient-specific patch sets or graphs. This motivates a more compact
representation that captures recurrent spatial organization at the cohort
level.

\begin{figure}[t]
\centering
\includegraphics[width=\linewidth]{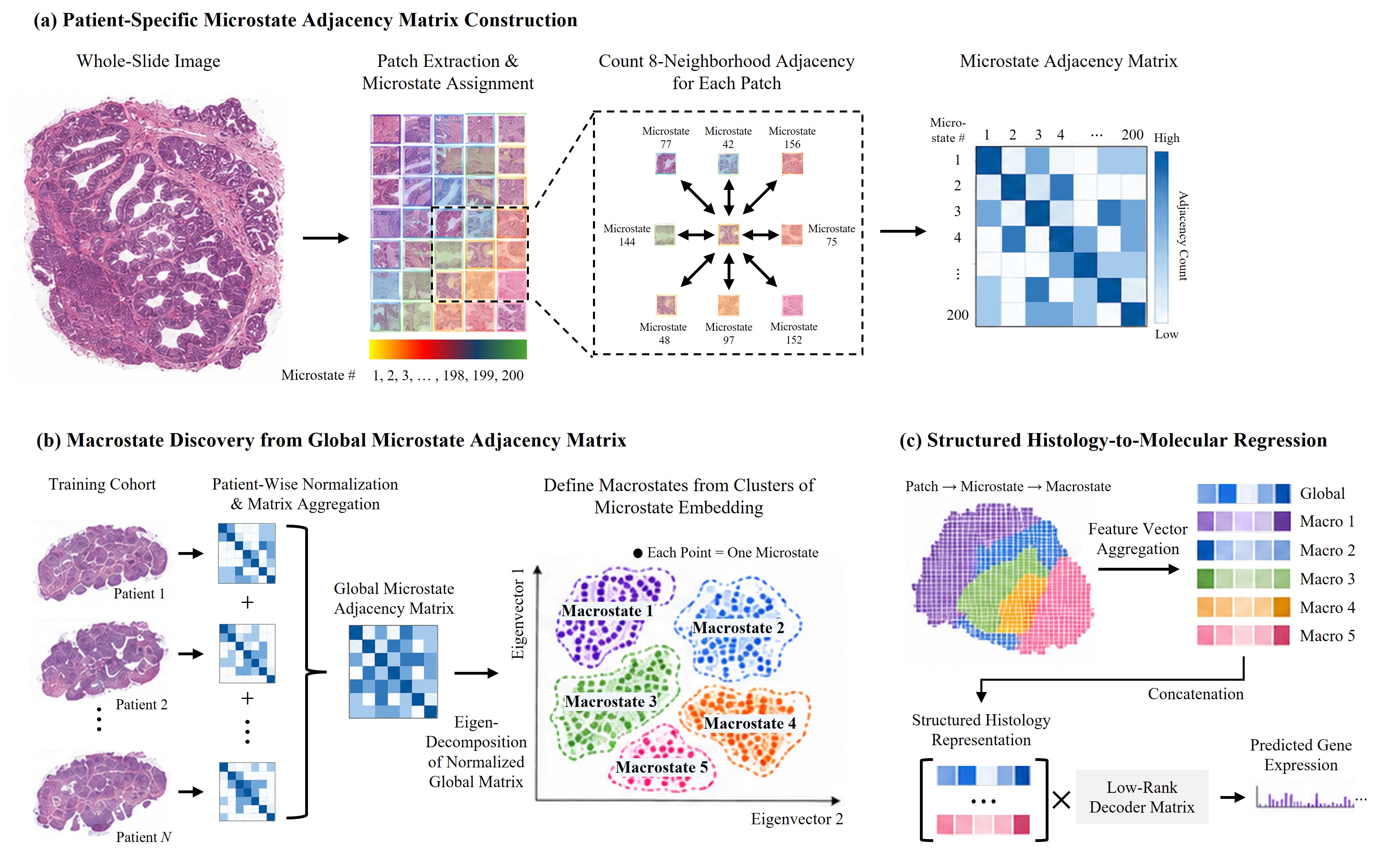}
\caption{Overview of MoSPR.}
\label{fig:overview}
\end{figure}

We propose \textbf{MoSPR} (\textbf{Mo}rpho-\textbf{S}patial
\textbf{P}rogram \textbf{R}egression), a linear framework that
compresses whole-slide histology into a compact \textit{morpho-spatial}
representation, defined here by the spatial adjacency of recurring
morphological states. To construct this representation, MoSPR first clusters
frozen patch embeddings into morphology microstates. For each
patient, spatial co-occurrences between neighboring patches are counted to
construct a microstate adjacency matrix (Fig.~\ref{fig:overview}(a)), which is normalized and aggregated
across the training cohort to form a global microstate adjacency matrix.
After row-normalizing this global matrix, eigendecomposition embeds
microstates according to their adjacency patterns, and clustering these embeddings defines a shared
set of macrostates (Fig.~\ref{fig:overview}(b)). Each slide is then represented
by a global morphology vector $M$ and abundance-weighted macrostate deviations
$S$, yielding $X=[M\mid S]$ (Fig.~\ref{fig:overview}(c)).


To reduce the dimensionality of molecular prediction, MoSPR learns a
low-rank expression basis $U_q\in\RR^{q\times G}$ from the training gene-expression targets ($Y_z\in\RR^{N\times G}$) and a regression matrix $W_q$
mapping $X$ to the corresponding molecular coefficients:
\begin{equation}
    \widehat{Y}_z = XW_qU_q,
    \qquad X=[M\mid S],
    \label{eq:mospr-main}
\end{equation}
where $q\ll G$. Thus, MoSPR combines morphology aggregation within
adjacency-informed states with regression to low-dimensional molecular
coefficients rather than directly to thousands of genes.

Across three cohorts from The Cancer Genome Atlas (TCGA; \citealp{weinstein2013cancer})---breast invasive
carcinoma (BRCA), kidney renal clear cell carcinoma (KIRC), and lung
adenocarcinoma (LUAD)---MoSPR achieves the highest mean gene-expression
prediction scores among all evaluated methods. Pathway scores derived from
its predicted expression profiles rank first in eight of nine comparisons
across Hallmark~\citep{liberzon2015hallmark}, Gene Ontology Biological Process (GO-BP)~\citep{geneontology2021}, and KEGG collections~\citep{kanehisa2000kegg}, without pathway-level supervision. Controlled ablations on BRCA show complementary improvements from the macrostate representation
and low-rank output model. On the same cohort, MoSPR trained with half of the
data exceeds the full-data gene-prediction score of the strongest competing baseline.
Finally, the linear formulation separates the predicted molecular profile
into global morphology and macrostate-specific contributions, supporting
qualitative interpretation of macrostate-associated molecular programs.

Our contributions are:
\begin{itemize}[leftmargin=*,itemsep=1pt,topsep=2pt]
\item a cohort-level morphology dictionary learned from patch adjacency,
yielding shared macrostates and a compact WSI representation without
spatial-omics supervision or large-scale patient-specific graph construction.

\item a linear framework coupling this representation with a
training-derived low-rank expression basis, enabling exact decomposition of each predicted molecular profile into global and macrostate-specific molecular contributions.



\item a systematic evaluation across three cancer cohorts, including
gene and pathway reconstruction, component ablations, training-set scaling, and
qualitative macrostate interpretation.


\end{itemize}

\section{Related Work}

Whole-slide gene-expression prediction from H\&E images has progressed
from patch-level aggregation~\citep{schmauch2020he2rna} to attention-,
transformer-, and clustering-based approaches~\citep{graziani2022abreg,
alsaafin2022trnaformer,senbabaoglu2024mosby,pizurica2024sequoia}.
General WSI aggregation methods further include attention-based
MIL~\citep{ilse2018abmil}, low-rank attention~\citep{xiang2023ilra},
structured state-space models~\citep{fillioux2023s4mil}, and Mamba-based
sequence models~\citep{yang2024mambamil,zhang2025twodmamba}.
CPNN~\citep{nishimura2026cpnn} additionally incorporates molecular
structure through single-cell-derived prototypes. MoSPR uses training-cohort patch adjacency to learn shared morphology states and couples state-wise aggregation with low-rank expression prediction.

Unlike spatial transcriptomics, which directly measures spatially resolved
molecular variation~\citep{stahl2016spatial}, MoSPR uses bulk molecular
supervision and does not reconstruct spatial expression maps. WSI methods
such as TransMIL~\citep{shao2021transmil} and
Patch-GCN~\citep{chen2021patchgcn} model contextual or spatial dependencies
over patient-specific patch sequences or graphs, whereas MoSPR constructs a cohort-level adjacency structure over shared morphology states and derives a compact macrostate representation.

\section{Methods}
\subsection{Patch feature extraction}
For patient $n$, let $\mathcal{X}^{(n)}=\{x_i^{(n)}\}_{i=1}^{N_n}$ denote the tissue patches and let $Y^{(n)}\in\RR^G$ denote the paired gene-expression vector. A frozen pathology encoder ($f_{enc}$) maps each patch to
\begin{equation}
    z_i^{(n)}=f_{\mathrm{enc}}(x_i^{(n)})\in\RR^{D_0},
\end{equation}
where $D_0=512$ for CONCH~\citep{lu2024conch}.

\subsection{Microstate-to-macrostate construction}
\paragraph{Microstate assignment and adjacency matrix construction.}
To discretize the continuous patch-embedding space into recurrent morphological patterns, we apply $k$-means clustering with $J=200$ clusters to a patient-balanced sample of training patch embeddings. We define each resulting cluster as a \emph{microstate}, represented by its centroid
\[
    \{\mu_j\}_{j=1}^{J}, \qquad \mu_j\in\RR^{D_0}.
\]
Each tissue patch is then assigned to the nearest microstate,
\begin{equation}
    a_i^{(n)}
    =
    \arg\min_{j\in\{1,\ldots,J\}}
    \left\|z_i^{(n)}-\mu_j\right\|_2^2,
\end{equation}
where $a_i^{(n)}\in\{1,\ldots,J\}$ denotes the microstate assigned to patch $i$ of patient $n$.

We then characterize how these microstates are spatially arranged within each WSI using 8-neighbor patch adjacency (Fig. ~\ref{fig:overview}(a)). For each pair of neighboring tissue patches $(p,q)$ assigned to microstates
$a_p^{(n)}=u$ and $a_q^{(n)}=v$, the corresponding entries of the patient-level \emph{microstate adjacency matrix}
$C^{(n)}\in\RR_+^{J\times J}$ are incremented:
\begin{equation}
    C_{uv}^{(n)} \leftarrow C_{uv}^{(n)}+1,
    \qquad
    C_{vu}^{(n)} \leftarrow C_{vu}^{(n)}+1.
\end{equation}
Thus, $C_{uv}^{(n)}$ quantifies how frequently patches belonging to microstates $u$ and $v$ occur as spatial neighbors within patient $n$, and $C^{(n)}$ is symmetric by construction.

To prevent patients with larger tissue areas from dominating the cohort-level
spatial structure, each patient-level microstate adjacency matrix is normalized
by its total adjacency mass:
\begin{equation}
    \widehat C^{(n)}
    =
    \frac{C^{(n)}}{\sum_{u,v}C_{uv}^{(n)}}.
\end{equation}
The global microstate adjacency matrix is then defined as the average of the
normalized patient-level matrices over the training cohort $\mathcal D_{\mathrm{tr}}$ (Fig. ~\ref{fig:overview}(b)):
\begin{equation}
    C_{\mathrm{global}}
    =
    \frac{1}{|\mathcal D_{\mathrm{tr}}|}
    \sum_{n\in\mathcal D_{\mathrm{tr}}}
    \widehat C^{(n)}.
    \label{eq:cglobal}
\end{equation}

This averaging gives equal weight to each training patient while preserving
symmetry and unit total mass ($C_{\mathrm{global}}=C_{\mathrm{global}}^\top$ and $\sum_{u,v}(C_{\mathrm{global}})_{uv}=1$).

\paragraph{Macrostate discovery.}
To define macrostates according to their spatial neighborhood patterns,
we first row-normalize the global microstate adjacency matrix. Let
\begin{equation}
    d_u=\sum_v(C_{\mathrm{global}})_{uv},
    \qquad
    P=\Delta^{-1}C_{\mathrm{global}},
    \qquad
    \Delta=\diag(d_1,\ldots,d_J).
\end{equation}
Here, $P\in\RR^{J\times J}$ and $P_{uv}$ represents the relative
frequency with which microstate $v$ occurs adjacent to microstate $u$.

We then compute the eigendecomposition of $P$,
\begin{equation}
    P\psi_\ell=\lambda_\ell\psi_\ell,
    \qquad
    \psi_\ell\in\RR^J.
\end{equation}
Following diffusion-map and spectral-clustering constructions
~\citep{coifman2006diffusion,ng2002spectral}, we exclude the trivial
stationary component associated with $\lambda_0=1$ and retain the next
$L=20$ nontrivial components. Each microstate $u$ is then represented by
the eigenvalue-weighted spectral embedding
\begin{equation}
    \phi_u=
    [\lambda_1\psi_1(u),\ldots,
    \lambda_L\psi_L(u)]
    \in\RR^{L}.
\end{equation}
The eigenvalue weighting gives greater influence to dominant spectral
components of the adjacency structure. Each embedding is subsequently
normalized to unit length,
\begin{equation}
    \bar\phi_u=
    \frac{\phi_u}{\|\phi_u\|_2+\varepsilon},
\end{equation}
where $\varepsilon=10^{-12}$ is a small numerical constant. Finally, $k$-means clustering is applied to the $J=200$
normalized spectral embeddings, grouping the microstates into $K=8$
macrostates. This clustering defines a fixed microstate-to-macrostate
mapping
\begin{equation}
    \rho:\{1,\ldots,J\}\rightarrow\{1,\ldots,K\},
\end{equation}
where $\rho(u)=k$ indicates that microstate $u$ is assigned to
macrostate $k$ (Fig. ~\ref{fig:overview}(b)).

\subsection{Morpho-spatial low-rank regression}
\paragraph{Structured histology representation.}
We represent each patient's WSI by combining global morphology with macrostate-specific deviations that summarize within-slide morphological heterogeneity across the learned states (Fig. ~\ref{fig:overview}(c)). The global morphology vector is the mean of the original patch features,
\begin{equation}
    M_n=\frac{1}{N_n}\sum_{i=1}^{N_n}z_i^{(n)}\in\RR^{D_0}.
    \label{eq:global-m}
\end{equation}
For macrostate $k$, let $\mathcal I_{nk}=\{i:\rho(a_i^{(n)})=k\}$, $N_{nk}=|\mathcal I_{nk}|$, and $p_{nk}=N_{nk}/N_n$. When $N_{nk}>0$, its mean patch feature and abundance-weighted deviation from the slide mean are
\begin{equation}
    h_{nk}=\frac{1}{N_{nk}}\sum_{i\in\mathcal I_{nk}}z_i^{(n)},
    \qquad
    s_{nk}=p_{nk}(h_{nk}-M_n)\in\RR^{D_0}.
    \label{eq:state-block}
\end{equation}
If macrostate $k$ is absent from patient $n$ ($N_{nk}=0$), we set $s_{nk}=0$. We concatenate the state blocks as $S_n=[s_{n1}\mid\cdots\mid s_{nK}]\in\RR^{KD_0}$ and form
\begin{equation}
    X_n=[M_n\mid S_n]\in\RR^{(K+1)D_0}.
    \label{eq:xms}
\end{equation}
The resulting vector $X_n$ serves as the structured histological representation of patient $n$. Stacking these representations across $N$ patients yields the matrix $X\in\RR^{N\times (K+1)D_0}$. 


\paragraph{Low-rank molecular factorization.}

Gene targets are standardized per gene using training-fold statistics,
yielding $Y_z\in\RR^{N\times G}$ (Supplementary Section~\ref{sec:supp-gene-preprocessing}). Rather than
predicting all $G$ genes independently, we exploit the strong correlation
structure of gene expression by learning a compact molecular basis directly
from the training cohort. PCA is fitted exclusively to the standardized
training-fold expression matrix, and the top $q$ principal directions define
\begin{equation}
    U_q\in\RR^{q\times G}.
\end{equation}
Each molecular profile is represented by a low-dimensional coefficient vector,
and the corresponding coefficient matrix is
\begin{equation}
    A_q
    =
    Y_zU_q^\top
    \in\RR^{N\times q}.
    \label{eq:molecular-coefficients}
\end{equation}
The resulting approximation
\begin{equation}
    Y_z\approx A_qU_q
\end{equation}
reduces the output regression problem from $G$ genes to $q\ll G$ molecular
coefficients.

\paragraph{Molecular regression and reconstruction.}

We fit a ridge regression matrix $W_q$ that maps the structured histology
representation $X=[M\mid S]$ to the low-dimensional molecular coefficient
space:
\begin{equation}
    W_q^*
    =
    \arg\min_{W_q}
    \|A_q-XW_q\|_F^2
    +
    \lambda_{\mathrm{ridge}}\|W_q\|_F^2,
    \label{eq:ridge-objective}
\end{equation}
where $W_q^*\in\RR^{(K+1)D_0\times q}$ is the fitted regression matrix.

The predicted molecular coefficients are projected back through the learned
basis to reconstruct the standardized gene-expression profile:
\begin{equation}
    \widehat Y_z
    =
    XW_q^*U_q.
    \label{eq:prediction-final}
\end{equation}
Intercept terms, which are included in the implementation, are omitted from
the notation for clarity.

Thus, the original high-dimensional histology-to-gene regression problem is
reduced to predicting only $q$ molecular coefficients from the
morpho-spatial representation. 






\subsection{Macrostate-level molecular interpretation}
\label{sec:macro-interp}
According to the block structure of $X=[M\mid S]$, the fitted regression
matrix can be partitioned as
\begin{equation}
    W_q^*
    =
    \begin{bmatrix}
        W_{q,\mathrm{global}}^* \\
        W_{q,1}^* \\
        \vdots \\
        W_{q,K}^*
    \end{bmatrix},
    \qquad
    W_{q,\mathrm{global}}^*,\,W_{q,k}^*
    \in\RR^{D_0\times q}.
\end{equation}
For patient $n$, the macrostate-$k$ contribution to the predicted
gene-expression profile is
\begin{equation}
    \Gamma_{nk}
    =
    s_{nk}W_{q,k}^*U_q
    \in\RR^G,
    \label{eq:macro-contribution}
\end{equation}
and the global morphology contribution is
\begin{equation}
    \Gamma_{n,\mathrm{global}}
    =
    M_n W_{q,\mathrm{global}}^*U_q
    \in\RR^G.
\end{equation}

Ignoring the intercept term for notational simplicity, the predicted profile
can therefore be decomposed as
\begin{equation}
    \widehat Y_{z,n}
    =
    \Gamma_{n,\mathrm{global}}
    +
    \sum_{k=1}^{K}\Gamma_{nk}.
    \label{eq:additive-decomposition}
\end{equation}
This provides an exact algebraic decomposition of the prediction into
contributions from global morphology and individual macrostates, without
requiring a post-hoc attribution method.


\section{Experimental Setup}

\subsection{Datasets and evaluation protocol}

We evaluate MoSPR on paired H\&E WSIs and bulk RNA-sequencing profiles from TCGA-BRCA, TCGA-KIRC, and TCGA-LUAD \citep{weinstein2013cancer}. Performance is assessed at two levels: gene-expression prediction and pathway-score reconstruction. We use four-fold patient-level cross-validation, where held-out patients in each outer fold are reserved exclusively for final evaluation and model fitting and hyperparameter selection are performed using only the corresponding training and validation partitions. Within each cancer type, all competing methods use identical patient-level splits and the same target-gene universe. Cohort statistics and fold-specific training, validation, and test sample counts are provided in Supplementary Table~\ref{tab:dataset-summary}.

\subsection{Evaluation metrics}


For gene-level evaluation, Pearson correlation coefficient (PCC) and Spearman rank correlation coefficient (SCC) are computed across held-out patients separately for each gene and then averaged across genes. For pathway-level evaluation, pathway scores are computed independently from the predicted and observed gene-expression profiles as the mean gene-wise z-score across the member genes of each pathway~\citep{lee2008pac}. SCC is then computed across held-out patients separately for each pathway. Performance within each pathway collection is summarized by the median SCC across pathways in the Hallmark~\citep{liberzon2015hallmark}, GO-BP~\citep{geneontology2021}, and KEGG~\citep{kanehisa2000kegg} collections.


\subsection{Benchmark methods}

We compare MoSPR with both general-purpose WSI aggregation methods and models developed specifically for transcriptomic prediction from histology. WSI aggregation baselines include max and mean pooling~\citep{wang2018revisiting}, ABMIL~\citep{ilse2018abmil}, ILRA~\citep{xiang2023ilra}, S4MIL~\citep{fillioux2023s4mil}, MambaMIL and SRMambaMIL~\citep{yang2024mambamil}, and 2DMamba~\citep{zhang2025twodmamba}. Transcriptomic prediction baselines include HE2RNA~\citep{schmauch2020he2rna}, AbReg~\citep{graziani2022abreg}, tRNAformer~\citep{alsaafin2022trnaformer}, MOSBY~\citep{senbabaoglu2024mosby}, SEQUOIA VIS~\citep{pizurica2024sequoia}, and CPNN~\citep{nishimura2026cpnn}.

\subsection{Implementation details}

Structural hyperparameters are fixed across all cohorts at $J=200$ microstates, $L=20$ spectral components, and $K=8$ macrostates. The molecular rank $q$ and ridge penalty $\lambda_{\mathrm{ridge}}$ are selected independently within each fold using the validation reconstruction error. Hyperparameter ranges and selection details are provided in Supplementary Section~\ref{sec:supp-hyper-select}.

\subsection{Additional analyses}


\paragraph{Ablation analysis.}
We isolate the contributions of the macrostate representation and low-rank
molecular prediction using matched model variants on TCGA-BRCA.
\emph{Spatial Low-Rank} corresponds to the full MoSPR formulation,
combining macrostate-based features with low-rank regression.
\emph{Spatial Direct} retains the same macrostate representation but replaces
the low-rank output model with direct gene-wise regression, whereas
\emph{Global Low-Rank} retains low-rank molecular prediction but removes the
macrostate-specific features. As a control, \emph{Shuffled-State Low-Rank} and \emph{Shuffled-State Direct} permute the learned microstate-to-macrostate mapping $\rho$ across microstates, preserving the
number and sizes of macrostates while disrupting their adjacency-derived
composition. All variants use the same patient-level
four-fold cross-validation splits as the main benchmarks
(Supplementary Table~\ref{tab:dataset-summary}). Performance is evaluated
using mean gene-level SCC and median Hallmark pathway SCC.

\paragraph{Data-efficiency analysis.}
To assess performance under limited paired histology--transcriptomic supervision, we conduct a training-set scaling analysis across the three cancer cohorts using the same patient-level four-fold splits as the main benchmarks (Supplementary Table~\ref{tab:dataset-summary}). Within each outer fold, the training and validation partitions are randomly subsampled at fractions $f\in\{0.1,0.25,0.5,0.75,1.0\}$, while the held-out test partition remains unchanged. For each fold and fraction, the sampled training and validation patients are fixed and shared across all competing methods. All data-derived representations and prediction models are recomputed using only the corresponding subsample. Performance is evaluated using mean gene-level SCC and median Hallmark pathway SCC on the unchanged test folds.



\paragraph{Macrostate interpretation.}
To qualitatively illustrate how learned macrostates map to tissue regions and
associate with molecular programs, we select a representative held-out patient
WSI from TCGA-BRCA and analyze it using the model trained in a single
cross-validation fold. We map the patch assignments of three selected macrostates onto
this WSI. For each macrostate, its model-derived gene-expression contribution
($\Gamma_{nk}$ in Section~\ref{sec:macro-interp}) is used for Hallmark pathway
enrichment analysis. The three pathways with the largest absolute enrichment
$z$-scores are reported for each macrostate.


\section{Results}

\subsection{MoSPR consistently improves cross-cancer gene-expression prediction}

Table~\ref{tab:gene-main} summarizes gene-level prediction performance.
MoSPR achieves the highest mean PCC and SCC in all six cohort--metric combinations
across BRCA, KIRC, and LUAD. Although the strongest competing baseline
varies across cohorts and metrics, MoSPR consistently exceeds the strongest competing baseline in each setting, with PCC/SCC gains of
$0.043/0.048$ in BRCA, $0.022/0.029$ in KIRC, and $0.039/0.040$ in LUAD.
The consistent gains across both linear and rank correlation indicate that
the improvement is not specific to a single correlation metric or cancer type.
The best- and worst-predicted individual genes are reported in
Supplementary Table~\ref{tab:supp-gene-best-worst}. In addition, MoSPR achieves these gains with the lowest parameter count ($0.96$M parameters) among the evaluated methods, while requiring only
$2.02$ inference GFLOPs after frozen CONCH feature extraction
(Supplementary Table~\ref{tab:supp-inference-cost}).

\begin{table}[h]
\caption{Cross-cancer gene-expression prediction. Entries report mean gene-wise PCC and SCC over four cross-validation folds. Best and second-best results are shown in bold and underlined, respectively. }
\label{tab:gene-main}
\centering
\begingroup
\small
\setlength{\tabcolsep}{2.3pt}
\renewcommand{\arraystretch}{0.85}

\begin{tabular}{lcccccc}
\toprule
& \multicolumn{2}{c}{BRCA} & \multicolumn{2}{c}{KIRC} & \multicolumn{2}{c}{LUAD}\\
\cmidrule(lr){2-3}\cmidrule(lr){4-5}\cmidrule(lr){6-7}
Method & PCC & SCC & PCC & SCC & PCC & SCC\\
\midrule
Max~\citep{wang2018revisiting} & 0.294 & 0.283 & 0.236 & 0.244 & 0.258 & 0.278\\
Mean~\citep{wang2018revisiting} & 0.336 & 0.338 & 0.285 & 0.293 & 0.292 & 0.315\\
ABMIL~\citep{ilse2018abmil} & \underline{0.370} & \underline{0.363} & 0.293 & 0.299 & 0.308 & 0.324\\
HE2RNA~\citep{schmauch2020he2rna} & 0.300 & 0.293 & 0.256 & 0.273 & 0.276 & 0.297\\
AbReg~\citep{graziani2022abreg} & 0.355 & 0.349 & 0.289 & 0.293 & 0.296 & 0.311\\
tRNAformer~\citep{alsaafin2022trnaformer} & 0.341 & 0.332 & \underline{0.312} & \underline{0.319} & 0.316 & \underline{0.336}\\
ILRA~\citep{xiang2023ilra} & 0.238 & 0.271 & 0.191 & 0.248 & 0.280 & 0.304\\
S4MIL~\citep{fillioux2023s4mil} & 0.328 & 0.330 & 0.269 & 0.274 & 0.299 & 0.315\\
MambaMIL~\citep{yang2024mambamil} & 0.363 & 0.356 & 0.290 & 0.290 & \underline{0.319} & 0.335\\
SRMambaMIL~\citep{yang2024mambamil} & 0.361 & 0.352 & 0.292 & 0.294 & 0.311 & 0.326\\
MOSBY~\citep{senbabaoglu2024mosby} & 0.343 & 0.349 & 0.264 & 0.289 & 0.289 & 0.308\\
SEQUOIA VIS~\citep{pizurica2024sequoia} & 0.353 & 0.337 & 0.310 & 0.314 & 0.317 & 0.331\\
2DMamba~\citep{zhang2025twodmamba} & 0.352 & 0.348 & 0.287 & 0.292 & 0.300 & 0.313\\
CPNN~\citep{nishimura2026cpnn} & 0.360 & 0.356 & 0.293 & 0.310 & 0.305 & 0.325\\
\midrule
\textbf{MoSPR (ours)} & \textbf{0.413} & \textbf{0.411} & \textbf{0.334} & \textbf{0.348} & \textbf{0.358} & \textbf{0.376}\\
\bottomrule
\end{tabular}
\endgroup
\end{table}

\subsection{Gene predictions preserve pathway-level functional variation}

MoSPR achieves the highest median pathway-wise SCC in eight of nine
comparisons across the three cohorts and the Hallmark, GO-BP, and KEGG
collections, and ranks second in the remaining comparison
(Table~\ref{tab:pathway-main}). Relative to the strongest competing method
in each setting, MoSPR improves median pathway-wise SCC by up to $0.051$,
with gains across all three pathway collections in BRCA and KIRC. Notably,
pathway scores are derived solely from predicted gene-expression profiles
and are not used for training. These results indicate that improvements in gene-expression prediction
extend to coordinated variation across biological programs.

\begin{table}[!t]
\caption{
Gene-derived pathway reconstruction across three TCGA cohorts. Entries report the median pathway-wise SCC within each collection, averaged over four cross-validation folds. Pathway scores are derived solely from predicted gene-expression profiles and are not used for training. Best and second-best results are shown in bold and underlined, respectively.
}
\label{tab:pathway-main}
\centering
\begingroup
\scriptsize
\setlength{\tabcolsep}{2.3pt}
\renewcommand{\arraystretch}{0.85}

\resizebox{\linewidth}{!}{%
\begin{tabular}{lccccccccc}
\toprule
& \multicolumn{3}{c}{BRCA} & \multicolumn{3}{c}{KIRC} & \multicolumn{3}{c}{LUAD}\\
\cmidrule(lr){2-4}\cmidrule(lr){5-7}\cmidrule(lr){8-10}
Method & Hallmark & GO-BP & KEGG & Hallmark & GO-BP & KEGG & Hallmark & GO-BP & KEGG\\
\midrule
Max~\citep{wang2018revisiting} & 0.356 & 0.337 & 0.321 & 0.281 & 0.284 & 0.309 & 0.420 & 0.379 & 0.388\\
Mean~\citep{wang2018revisiting} & 0.469 & 0.448 & 0.415 & 0.367 & 0.379 & 0.415 & 0.498 & 0.446 & 0.452\\
ABMIL~\citep{ilse2018abmil} & \underline{0.500} & \underline{0.476} & \underline{0.446} & 0.378 & 0.388 & 0.418 & 0.519 & 0.465 & 0.474\\
HE2RNA~\citep{schmauch2020he2rna} & 0.333 & 0.329 & 0.300 & 0.299 & 0.321 & 0.340 & 0.388 & 0.349 & 0.365\\
AbReg~\citep{graziani2022abreg} & 0.481 & 0.453 & 0.425 & \underline{0.397} & \underline{0.389} & \underline{0.420} & 0.507 & 0.443 & 0.462\\
tRNAformer~\citep{alsaafin2022trnaformer} & 0.412 & 0.381 & 0.345 & 0.367 & 0.368 & 0.404 & 0.467 & 0.415 & 0.424\\
ILRA~\citep{xiang2023ilra} & 0.385 & 0.358 & 0.330 & 0.301 & 0.317 & 0.340 & 0.463 & 0.418 & 0.430\\
S4MIL~\citep{fillioux2023s4mil} & 0.434 & 0.414 & 0.383 & 0.301 & 0.329 & 0.343 & 0.481 & 0.428 & 0.436\\
MambaMIL~\citep{yang2024mambamil} & 0.479 & 0.458 & 0.423 & 0.308 & 0.326 & 0.345 & \textbf{0.529} & \underline{0.481} & \underline{0.487}\\
SRMambaMIL~\citep{yang2024mambamil} & 0.463 & 0.448 & 0.411 & 0.306 & 0.315 & 0.338 & 0.514 & 0.455 & 0.472\\
MOSBY~\citep{senbabaoglu2024mosby} & 0.400 & 0.381 & 0.354 & 0.257 & 0.263 & 0.293 & 0.348 & 0.312 & 0.324\\
SEQUOIA VIS~\citep{pizurica2024sequoia} & 0.444 & 0.413 & 0.385 & 0.384 & 0.388 & 0.420 & 0.488 & 0.441 & 0.450\\
2DMamba~\citep{zhang2025twodmamba} & 0.460 & 0.439 & 0.408 & 0.381 & 0.370 & 0.388 & 0.501 & 0.443 & 0.457\\
CPNN~\citep{nishimura2026cpnn} & 0.391 & 0.377 & 0.327 & 0.282 & 0.293 & 0.349 & 0.427 & 0.380 & 0.388\\
\midrule
\textbf{MoSPR (ours)} & \textbf{0.551} & \textbf{0.520} & \textbf{0.486} & \textbf{0.402} & \textbf{0.415} & \textbf{0.447} & \underline{0.528} & \textbf{0.482} & \textbf{0.491}\\
\bottomrule
\end{tabular}
}

\endgroup
\end{table}

\subsection{Ablation of spatial and low-rank structure}

On TCGA-BRCA, Global Low-Rank improves mean gene SCC from $0.348$
(Global Direct) to $0.380$, while Spatial Direct reaches $0.395$
(Table~\ref{tab:ablation}). Combining the macrostate representation with
low-rank regression (Spatial Low-Rank; MoSPR) yields the highest mean gene SCC ($0.411$).
Importantly, applying baseline-specific low-rank output controls to ABMIL
and tRNAformer does not reproduce MoSPR's performance
(Supplementary Table~\ref{tab:supp-lowrank-control}), suggesting that
low-rank output compression alone is insufficient to explain the gain.
Randomizing the microstate-to-macrostate assignments reduces performance
to $0.360$ for Shuffled-State Direct and $0.379$ for Shuffled-State
Low-Rank, both below their learned-macrostate counterparts ($0.395$ and
$0.411$, respectively). For Hallmark reconstruction, the shuffled-state variants both
reach $0.499$, whereas Spatial Direct ($0.552$) and MoSPR ($0.551$) perform
similarly and both exceed Global Low-Rank ($0.481$). 

Beyond predictive performance, the learned macrostate organization was also
stable under patient resampling. Across 300 patient-level bootstrap resamples
per cohort with fixed microstates, mean Adjusted Rand Index (ARI) values  relative to the corresponding
full-cohort partitions were $0.769$, $0.653$, and $0.786$ for TCGA-BRCA,
TCGA-KIRC, and TCGA-LUAD, respectively
(Supplementary Table~\ref{tab:supp-macrostate-ari}).
Together, these results indicate complementary contributions of the macrostate
representation and low-rank output model to gene prediction, with pathway-level
gains primarily associated with the learned macrostate representation.

\begin{table}[ht]
\caption{
Component ablation on TCGA-BRCA. Best and second-best results are shown in bold and underlined, respectively.}
\label{tab:ablation}
\centering
\scriptsize
\setlength{\tabcolsep}{4.0pt}
\renewcommand{\arraystretch}{0.95}

\begin{tabular}{lcccc}
\toprule
Variant & Histology Input& Molecular Output& Gene SCC & Hallmark SCC\\
\midrule
Global Direct & $M$ & Direct & 0.348 & 0.458\\
Global Low-Rank & $M$ & $U_q$ & 0.380 & 0.481\\
Shuffled-State Direct & $[M\mid S_{\mathrm{shuffle}}]$ & Direct & 0.360 & 0.499\\
Shuffled-State Low-Rank & $[M\mid S_{\mathrm{shuffle}}]$ & $U_q$ & 0.379 & 0.499\\
Spatial Direct & $[M\mid S]$ & Direct & \underline{0.395} & \textbf{0.552}\\
\textbf{Spatial Low-Rank (MoSPR)} & $[M\mid S]$ & $U_q$ & \textbf{0.411} & \underline{0.551}\\
\bottomrule
\end{tabular}
\end{table}


\subsection{Data efficiency under limited paired supervision}

\begin{figure}[ht]
\centering
\includegraphics[width=0.8\linewidth]{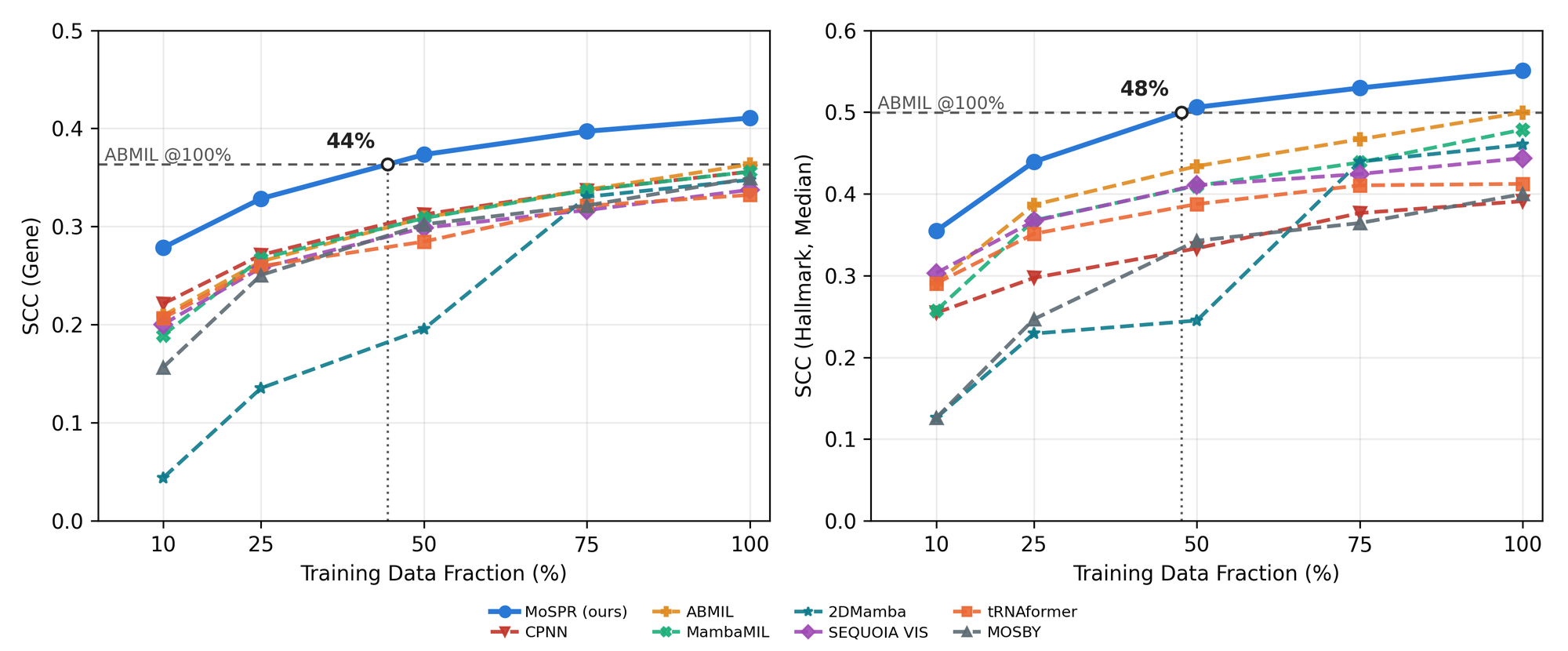}
\caption{Training-set scaling on TCGA-BRCA.}
\label{fig:data-efficiency}
\end{figure}

Across reduced training-set sizes, MoSPR consistently preserves its performance advantage, including against ABMIL, the strongest competing baseline on TCGA-BRCA (Fig. \ref{fig:data-efficiency}). For gene-expression prediction, MoSPR reaches the full-data ABMIL performance using only approximately $44\%$ of the available training data. Similarly, for Hallmark pathway reconstruction, MoSPR trained with approximately $48\%$ of the data matches the performance of ABMIL trained with $100\%$. These results indicate that MoSPR can recover the performance of the strongest full-data baseline with roughly half of the paired histology--transcriptomic supervision. Corresponding training-set scaling results for TCGA-KIRC and TCGA-LUAD are provided in Supplementary Figs.~\ref{fig:supp-data-efficiency-kirc} and~\ref{fig:supp-data-efficiency-luad}.

\subsection{Qualitative characterization of morpho-spatial macrostates}

\begin{figure*}[ht]
\centering
\includegraphics[width=0.85\linewidth]{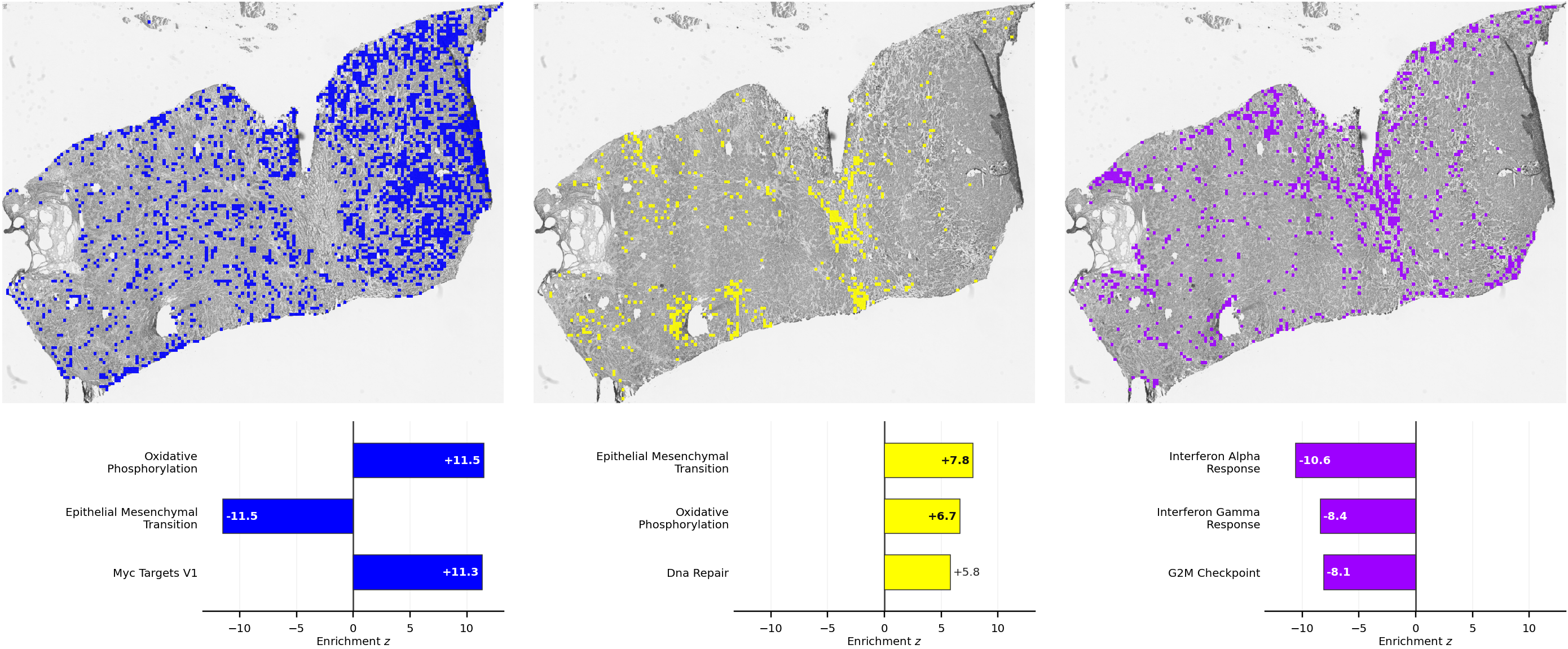}
\caption{Representative macrostates and their associated Hallmark pathway enrichments.}
\label{fig:macro-biology}
\end{figure*}

Three representative macrostates localize to distinct, spatially coherent regions in a representative WSI (Figure~\ref{fig:macro-biology}). Their corresponding gene-expression contributions show distinct Hallmark enrichment patterns, including oxidative phosphorylation, epithelial--mesenchymal transition, and interferon gamma response. This qualitative example suggests that the learned macrostates capture morphologically distinct tissue regions associated with different molecular programs.


\section{Discussion and Conclusion}

MoSPR couples a shared morphology dictionary informed by training-cohort
adjacency with low-rank gene-expression regression, summarizing within-slide
morphological heterogeneity across learned macrostates while reducing the
dimension of supervised regression. The gene- and pathway-level results,
together with the ablation and training-set scaling analyses, support this
combination for bulk expression prediction. The fitted linear formulation
further permits model-based interpretation of global and macrostate-specific
molecular contributions.

Several limitations remain. Macrostates summarize recurrent local adjacency
patterns rather than complete two-dimensional tissue geometry, and bulk RNA
measurements cannot validate whether macrostate-associated molecular programs
are spatially localized to the corresponding tissue regions. Accordingly, the
biological characterization remains qualitative. Evaluation within TCGA does
not establish generalization to independent cohorts or institutions, and
external cohorts and spatially resolved molecular measurements will be
important for validating both predictive robustness and the biological
interpretation of the learned macrostates.


\newpage

\section*{AI Use Statement}

Generative AI tools were used for translation, LaTeX formatting and conversion
of manuscript elements such as tables and figures, grammatical proofreading,
and code refactoring. Generative AI tools were not used to generate experimental
measurements, fabricate numerical results, or determine the reported scientific
findings. All AI-assisted text, code, analyses, and manuscript components were
reviewed and verified by the authors, who take full responsibility for the final
content of this manuscript.

\section*{Ethics Statement}

This study uses de-identified, publicly available data from The Cancer Genome
Atlas (TCGA), including the TCGA-BRCA, TCGA-KIRC, and TCGA-LUAD cohorts, and
does not involve new participant recruitment or the collection of new clinical
specimens. The study is intended for research and hypothesis generation rather
than direct clinical decision-making. Potential sources of bias include cohort
composition, tissue processing, scanner and staining differences, and
under-representation of demographic groups. All TCGA data were used in
accordance with the applicable data-access and data-use requirements.

\section*{Reproducibility Statement}

An implementation of MoSPR is available at the repository linked in the abstract. The
repository includes the code required to reproduce the proposed method and main
experimental analyses, together with patient-level data splits, preprocessing
procedures, model configurations, and scripts for generating the reported tables
and figures. Fold-specific model selection and evaluation procedures are
described in the manuscript and supplementary material. Data-derived quantities
are constructed independently within each training fold to avoid information
leakage. Where redistribution of derived data or intermediate artifacts is
restricted by the terms of the original datasets, scripts for deterministic
regeneration are provided where possible.

\newpage

\bibliography{references}

@article{schmauch2020he2rna,
  title={A deep learning model to predict RNA-Seq expression of tumours from whole slide images},
  author={Schmauch, Beno{\^\i}t and Romagnoni, Alberto and Pronier, Elodie and Saillard, Charlie and Maill{\'e}, Pascale and Calderaro, Julien and Kamoun, Aur{\'e}lie and Sefta, Meriem and Toldo, Sylvain and Zaslavskiy, Mikhail and others},
  journal={Nature communications},
  volume={11},
  number={1},
  pages={3877},
  year={2020},
  publisher={Nature Publishing Group UK London}
}

@article{alsaafin2022trnaformer,
  title={Learning to predict RNA sequence expressions from whole slide images with applications for search and classification},
  author={Alsaafin, Areej and Safarpoor, Amir and Sikaroudi, Milad and Hipp, Jason D and Tizhoosh, Hamid R},
  journal={Communications Biology},
  volume={6},
  number={1},
  pages={304},
  year={2023},
  publisher={Nature Publishing Group UK London}
}

@article{senbabaoglu2024mosby,
  title={MOSBY enables multi-omic inference and spatial biomarker discovery from whole slide images},
  author={{\c{S}}enbabao{\u{g}}lu, Yasin and Prabhakar, Vignesh and Khormali, Aminollah and Eastham, Jeff and Liu, Evan and Warner, Elisa and Nabet, Barzin and Srivastava, Minu and Ballinger, Marcus and Liu, Kai},
  journal={Scientific Reports},
  volume={14},
  number={1},
  pages={18271},
  year={2024},
  publisher={Nature Publishing Group UK London}
}

@article{pizurica2024sequoia,
  title={Digital profiling of gene expression from histology images with linearized attention},
  author={Pizurica, Marija and Zheng, Yuanning and Carrillo-Perez, Francisco and Noor, Humaira and Yao, Wei and Wohlfart, Christian and Vladimirova, Antoaneta and Marchal, Kathleen and Gevaert, Olivier},
  journal={Nature Communications},
  volume={15},
  number={1},
  pages={9886},
  year={2024},
  publisher={Nature Publishing Group UK London}
}

@inproceedings{nishimura2026cpnn,
  title={Cell-Type Prototype-Informed Neural Network for Gene Expression Estimation from Pathology Images},
  author={Nishimura, Kazuya and Bise, Ryoma and Matsuo, Shinnosuke and Hirose, Haruka and Kojima, Yasuhiro},
  booktitle={Proceedings of the IEEE/CVF Conference on Computer Vision and Pattern Recognition},
  year={2026}
}

@inproceedings{graziani2022abreg,
  title={Attention-based interpretable regression of gene expression in histology},
  author={Graziani, Mara and Marini, Niccolo and Deutschmann, Nicolas and Janakarajan, Nikita and M{\"u}ller, Henning and Rodr{\'i}guez Mart{\'i}nez, Mar{\'i}a},
  booktitle={MICCAI Workshop on Computational Pathology},
  pages={44--60},
  year={2022},
  organization={Springer}
}

@article{lu2024conch,
  title={A visual-language foundation model for computational pathology},
  author={Lu, Ming Y and Chen, Bowen and Williamson, Drew FK and Chen, Richard J and Liang, Ivy and Ding, Tong and Jaume, Guillaume and Odintsov, Igor and Le, Long Phi and Gerber, Georg and others},
  journal={Nature Medicine},
  volume={30},
  number={3},
  pages={863--874},
  year={2024}
}

@inproceedings{ng2002spectral,
  title={On spectral clustering: Analysis and an algorithm},
  author={Ng, Andrew Y and Jordan, Michael I and Weiss, Yair},
  booktitle={Advances in Neural Information Processing Systems},
  volume={14},
  year={2002}
}

@article{coifman2006diffusion,
  title={Diffusion maps},
  author={Coifman, Ronald R and Lafon, St{\'e}phane},
  journal={Applied and Computational Harmonic Analysis},
  volume={21},
  number={1},
  pages={5--30},
  year={2006},
  doi={10.1016/j.acha.2006.04.006}
}

@article{liberzon2015hallmark,
  title={The molecular signatures database hallmark gene set collection},
  author={Liberzon, Arthur and Birger, Chet and Thorvaldsdottir, Helga and Ghandi, Mahmoud and Mesirov, Jill P and Tamayo, Pablo},
  journal={Cell Systems},
  volume={1},
  number={6},
  pages={417--425},
  year={2015}
}

@article{geneontology2021,
  title={The Gene Ontology resource: enriching a {GOld} mine},
  author={{The Gene Ontology Consortium}},
  journal={Nucleic Acids Research},
  volume={49},
  number={D1},
  pages={D325--D334},
  year={2021}
}

@article{kanehisa2000kegg,
  title={{KEGG}: Kyoto Encyclopedia of Genes and Genomes},
  author={Kanehisa, Minoru and Goto, Susumu},
  journal={Nucleic Acids Research},
  volume={28},
  number={1},
  pages={27--30},
  year={2000}
}

@article{wang2018revisiting,
  title={Revisiting multiple instance neural networks},
  author={Wang, Xinggang and Yan, Yongluan and Tang, Peng and Bai, Xiang and Liu, Wenyu},
  journal={Pattern Recognition},
  volume={74},
  pages={15--24},
  year={2018}
}

@article{lee2008pac,
  title={Inferring pathway activity toward precise disease classification},
  author={Lee, Eunjung and Chuang, Han-Yu and Kim, Jong-Won and Ideker, Trey and Lee, Doheon},
  journal={PLOS Computational Biology},
  volume={4},
  number={11},
  pages={e1000217},
  year={2008}
}

@article{coudray2018classification,
  title={Classification and mutation prediction from non--small cell lung cancer histopathology images using deep learning},
  author={Coudray, Nicolas and Ocampo, Paolo Santiago and Sakellaropoulos, Theodore and Narula, Navneet and Snuderl, Matija and Feny{\"o}, David and Moreira, Andre L and Razavian, Narges and Tsirigos, Aristotelis},
  journal={Nature Medicine},
  volume={24},
  number={10},
  pages={1559--1567},
  year={2018}
}

@article{saillard2023msintuit,
  title={Validation of {MSIntuit} as an {AI}-based pre-screening tool for {MSI} detection from colorectal cancer histology slides},
  author={Saillard, Charlie and Dubois, R{\'e}my and Tchita, Oussama and Loiseau, Nicolas and Garcia, Thierry and Adriansen, Aur{\'e}lie and Carpentier, S{\'e}verine and Reyre, Joelle and Enea, Diana and von Loga, Katharina and Kamoun, Aur{\'e}lie and Rossat, St{\'e}phane and Wiscart, Corentin and Sefta, Meriem and Auffret, Micha{\"e}l and Guillou, Lionel and Fouillet, Arnaud and Kather, Jakob Nikolas and Svrcek, Magali},
  journal={Nature Communications},
  volume={14},
  number={1},
  pages={6695},
  year={2023}
}

@inproceedings{ilse2018abmil,
  title={Attention-based deep multiple instance learning},
  author={Ilse, Maximilian and Tomczak, Jakub and Welling, Max},
  booktitle={International Conference on Machine Learning},
  pages={2127--2136},
  year={2018},
  organization={PMLR}
}

@inproceedings{xiang2023ilra,
  title={Exploring low-rank property in multiple instance learning for whole slide image classification},
  author={Xiang, Jinxi and Zhang, Jun},
  booktitle={International Conference on Learning Representations},
  year={2023}
}

@inproceedings{fillioux2023s4mil,
  title={Structured state space models for multiple instance learning in digital pathology},
  author={Fillioux, Leo and Boyd, Joseph and Vakalopoulou, Maria and Cournede, Paul-Henry and Christodoulidis, Stergios},
  booktitle={Medical Image Computing and Computer Assisted Intervention},
  pages={594--604},
  year={2023},
  organization={Springer}
}

@inproceedings{yang2024mambamil,
  title={{MambaMIL}: Enhancing long sequence modeling with sequence reordering in computational pathology},
  author={Yang, Shu and Wang, Yihui and Chen, Hao},
  booktitle={Medical Image Computing and Computer Assisted Intervention},
  pages={296--306},
  year={2024},
  organization={Springer}
}

@inproceedings{zhang2025twodmamba,
  title={{2DMamba}: Efficient state space model for image representation with applications on giga-pixel whole slide image classification},
  author={Zhang, Jingwei and Nguyen, Anh Tien and Han, Xi and Trinh, Vincent Quoc-Huy and Qin, Hong and Samaras, Dimitris and Hosseini, Mahdi S},
  booktitle={Proceedings of the IEEE/CVF Conference on Computer Vision and Pattern Recognition},
  pages={3583--3592},
  year={2025}
}

@article{stahl2016spatial,
  title   = {Visualization and analysis of gene expression in tissue sections by spatial transcriptomics},
  author  = {St{\aa}hl, Patrik L. and Salm{\'e}n, Fredrik and Vickovic, Sanja and Lundmark, Anna and Navarro, Jos{\'e} Fern{\'a}ndez and Magnusson, Jens and Giacomello, Stefania and Asp, Michaela and Westholm, Jakub O. and Huss, Mikael and Mollbrink, Annelie and Linnarsson, Sten and Codeluppi, Simone and Borg, {\AA}ke and Pont{\'e}n, Fredrik and Costea, Paul I. and Sahl{\'e}n, Pelin and Mulder, Jan and Bergmann, Olaf and Lundeberg, Joakim and Fris{\'e}n, Jonas},
  journal = {Science},
  volume  = {353},
  number  = {6294},
  pages   = {78--82},
  year    = {2016},
  doi     = {10.1126/science.aaf2403}
}

@inproceedings{shao2021transmil,
  title     = {TransMIL: Transformer Based Correlated Multiple Instance Learning for Whole Slide Image Classification},
  author    = {Shao, Zhuchen and Bian, Hao and Chen, Yang and Wang, Yifeng and Zhang, Jian and Ji, Xiangyang and Zhang, Yongbing},
  booktitle = {Advances in Neural Information Processing Systems},
  volume    = {34},
  pages     = {2136--2147},
  year      = {2021}
}

@inproceedings{chen2021patchgcn,
  title     = {Whole Slide Images are 2D Point Clouds: Context-Aware Survival Prediction using Patch-Based Graph Convolutional Networks},
  author    = {Chen, Richard J. and Lu, Ming Y. and Wang, Jingwen and Williamson, Drew F. K. and Rodig, Scott J. and Lindeman, Neal I. and Mahmood, Faisal},
  booktitle = {Medical Image Computing and Computer Assisted Intervention -- MICCAI 2021},
  pages     = {339--349},
  year      = {2021},
  doi       = {10.1007/978-3-030-87237-3_33}
}

@article{weinstein2013cancer,
  title={The cancer genome atlas pan-cancer analysis project},
  author={Weinstein, John N and Collisson, Eric A and Mills, Gordon B and Shaw, Kenna R and Ozenberger, Brad A and Ellrott, Kyle and Shmulevich, Ilya and Sander, Chris and Stuart, Joshua M},
  journal={Nature genetics},
  volume={45},
  number={10},
  pages={1113--1120},
  year={2013},
  publisher={Nature Publishing Group}
}

@article{hubert1985comparing,
  title={Comparing partitions},
  author={Hubert, Lawrence and Arabie, Phipps},
  journal={Journal of classification},
  volume={2},
  number={1},
  pages={193--218},
  year={1985},
  publisher={Springer}
}
\bibliographystyle{plainnat}

\newpage

\appendix

\newpage

\section{Supplementary Methods}

\subsection{Gene-expression preprocessing}
\label{sec:supp-gene-preprocessing}

Let $T\in\RR_+^{N\times G}$ denote the TPM expression matrix restricted
to the $G$ target genes used for evaluation. For each sample, the
target-gene TPM profile was normalized within this gene subset to a total
of $10^4$ and log-transformed:
\begin{equation}
    Y_{ng}
    =
    \log\left(
        1+
        10^4
        \frac{T_{ng}}
        {\sum_{g'=1}^{G}T_{ng'}}
    \right).
    \label{eq:supp-expression-preprocessing}
\end{equation}
Thus, $Y\in\RR^{N\times G}$ represents log1p-transformed CP10K
expression within the target-gene subset.

Within each cross-validation fold, these values were further standardized
gene-wise using statistics estimated exclusively from the training
partition. Let $\mu_{\mathrm{tr},g}$ and $\sigma_{\mathrm{tr},g}$ denote
the training-fold mean and standard deviation of gene $g$. The regression
targets were
\begin{equation}
    (Y_z)_{ng}
    =
    \frac{
        Y_{ng}-\mu_{\mathrm{tr},g}
    }{
        \sigma_{\mathrm{tr},g}
    }.
    \label{eq:supp-target-standardization}
\end{equation}
The same training-fold mean and standard deviation were applied to the
validation and test partitions.

\subsection{Hyperparameter selection}
\label{sec:supp-hyper-select}

The structural hyperparameters were fixed across cohorts at
$J=200$ microstates, $K=8$ macrostates, and $L=20$ spectral components.
Additional searches over $K$ and $L$ did not improve
validation performance and frequently selected less stable macrostate
partitions. We therefore retained the fixed structural configuration
across all cohorts.

The molecular rank $q$ and ridge penalty $\lambda_{\mathrm{ridge}}$ were
selected independently within each fold from
\begin{equation}
    q\in\{4,8,16,32,48,64,96,128\},
    \qquad
    \lambda_{\mathrm{ridge}}
    \in\{1,10,10^2,10^3,10^4,10^5\}.
\end{equation}
For each candidate pair $(q,\lambda_{\mathrm{ridge}})$, model selection
was based on the mean squared reconstruction error over the validation
samples and genes:
\begin{equation}
    \mathcal L_{\mathrm{val}}
    =
    \frac{1}{N_{\mathrm{val}}G}
    \left\|
        \widehat Y_{z,\mathrm{val}}
        -
        Y_{z,\mathrm{val}}
    \right\|_F^2.
\end{equation}


After selecting $(q,\lambda_{\mathrm{ridge}})$, MoSPR and its fold-dependent
transformations were refitted on the combined training and validation
partitions before a single evaluation on the held-out test partition.


\clearpage

\section{Supplementary Tables}

\setcounter{table}{0}
\renewcommand{\thetable}{S\arabic{table}}
\renewcommand{\theHtable}{S\arabic{table}}

\begin{table}[!h]
\caption{
Dataset summary and fold-specific sample counts used for four-fold
cross-validation.
}
\label{tab:dataset-summary}
\centering
\small
\setlength{\tabcolsep}{4pt}
\renewcommand{\arraystretch}{1.0}

\resizebox{\linewidth}{!}{%
\begin{tabular}{lrrrcrrrrrr}
\toprule
&
\multicolumn{3}{c}{Cohort Summary}
&
&
\multicolumn{2}{c}{Train}
&
\multicolumn{2}{c}{Validation}
&
\multicolumn{2}{c}{Test}
\\
\cmidrule(lr){2-4}
\cmidrule(lr){6-7}
\cmidrule(lr){8-9}
\cmidrule(lr){10-11}

Cohort
& Patients
& Slides
& Genes
& Fold
& Patients & Slides
& Patients & Slides
& Patients & Slides
\\
\midrule

\multirow{4}{*}{TCGA-BRCA}
& \multirow{4}{*}{1{,}037}
& \multirow{4}{*}{1{,}467}
& \multirow{4}{*}{14{,}042}
& 0 & 519 & 734 & 259 & 366 & 259 & 367 \\
& & & & 1 & 519 & 733 & 259 & 367 & 259 & 367 \\
& & & & 2 & 518 & 733 & 259 & 367 & 260 & 367 \\
& & & & 3 & 518 & 734 & 260 & 367 & 259 & 366 \\
\midrule
\multirow{4}{*}{TCGA-KIRC}
& \multirow{4}{*}{342}
& \multirow{4}{*}{681}
& \multirow{4}{*}{14{,}295}
& 0 & 170 & 340 & 86 & 170 & 86 & 171 \\
& & & & 1 & 171 & 340 & 86 & 171 & 85 & 170 \\
& & & & 2 & 172 & 341 & 85 & 170 & 85 & 170 \\
& & & & 3 & 171 & 341 & 85 & 170 & 86 & 170 \\
\midrule
\multirow{4}{*}{TCGA-LUAD}
& \multirow{4}{*}{482}
& \multirow{4}{*}{756}
& \multirow{4}{*}{14{,}514}
& 0 & 241 & 378 & 121 & 189 & 120 & 189 \\
& & & & 1 & 242 & 378 & 120 & 189 & 120 & 189 \\
& & & & 2 & 241 & 378 & 120 & 189 & 121 & 189 \\
& & & & 3 & 240 & 378 & 121 & 189 & 121 & 189 \\

\bottomrule
\end{tabular}
}
\end{table}

\begin{table}[!h]
\caption{
Per-gene MoSPR performance on TCGA-BRCA, TCGA-KIRC, and TCGA-LUAD.
Shown are the 10 best- and worst-predicted genes ranked by mean
SCC across four cross-validation folds.
}
\label{tab:supp-gene-best-worst}
\centering
\small
\setlength{\tabcolsep}{5pt}
\renewcommand{\arraystretch}{1.0}

\begin{tabular}{lcccccc}
\toprule
& \multicolumn{2}{c}{BRCA} & \multicolumn{2}{c}{KIRC} & \multicolumn{2}{c}{LUAD} \\
\cmidrule(lr){2-3}\cmidrule(lr){4-5}\cmidrule(lr){6-7}
Rank & Gene & SCC & Gene & SCC & Gene & SCC \\
\midrule
\multicolumn{7}{l}{\textit{Best-predicted}} \\
1  & PODN      & 0.781 & ACAA2      & 0.728 & TPX2       & 0.750 \\
2  & HTRA1     & 0.762 & LINC01507  & 0.719 & PRR11      & 0.742 \\
3  & ZCCHC24   & 0.759 & EMX2OS     & 0.715 & CENPA      & 0.736 \\
4  & COL14A1   & 0.759 & LDB2       & 0.715 & CCNA2      & 0.736 \\
5  & LRRC17    & 0.751 & AGTR1      & 0.713 & TROAP      & 0.735 \\
6  & MFAP4     & 0.751 & TMEM204    & 0.709 & NCAPH      & 0.734 \\
7  & SPARCL1   & 0.751 & PITPNC1    & 0.708 & KIF4A      & 0.734 \\
8  & GLT8D2    & 0.749 & ERG        & 0.707 & BIRC5      & 0.730 \\
9  & COL1A2    & 0.747 & FRMD3      & 0.704 & CDCA5      & 0.729 \\
10 & SFRP2     & 0.742 & ESAM       & 0.703 & KPNA2      & 0.727 \\
\midrule
\multicolumn{7}{l}{\textit{Worst-predicted}} \\
1  & LINC01287 & -0.124 & AC093326.1 & -0.120 & LILRA2     & -0.167 \\
2  & MAEL      & -0.072 & PLIN5      & -0.101 & NEFL       & -0.083 \\
3  & PCDH9     & -0.063 & NEURL1     & -0.098 & HIST1H1B   & -0.074 \\
4  & MRPL53    & -0.061 & ZFR2       & -0.082 & KCNE5      & -0.051 \\
5  & PRSS21    & -0.055 & HIST1H3F   & -0.081 & HIST2H2AB  & -0.042 \\
6  & DLK1      & -0.054 & HIST2H2AB  & -0.081 & LATS2-AS1  & -0.036 \\
7  & HIST1H4A  & -0.033 & RNF212     & -0.076 & ADCYAP1    & -0.035 \\
8  & PCSK1     & -0.028 & ACSBG1     & -0.075 & L1TD1      & -0.035 \\
9  & HOXB8     & -0.027 & HIST1H2AJ  & -0.074 & CADM2      & -0.035 \\
10 & LINC00221 & -0.024 & CAGE1      & -0.073 & HIST1H1E   & -0.031 \\
\bottomrule
\end{tabular}
\end{table}

\begin{table}[!h]
\caption{
Model size and inference complexity on TCGA-BRCA after frozen CONCH feature
extraction. Parameter counts exclude the shared CONCH encoder and buffers such as
standardization statistics. The MoSPR count includes the microstate centroids and the PCA basis, which are fixed
during training and required at inference. GFLOPs are computed using the cohort
median of $9{,}609$ patches for full-bag methods; HE2RNA, tRNAformer, and SEQUOIA VIS
use their architecture-specific clustered inputs of 100, 49, and 100 super-patches,
respectively.
}
\label{tab:supp-inference-cost}
\centering
\small
\setlength{\tabcolsep}{6pt}
\renewcommand{\arraystretch}{1.0}

\begin{tabular}{lrr}
\toprule
Method & \# Params & GFLOPs \\
\midrule
Max              & 7.47M  & 5.14   \\
Mean             & 7.47M  & 5.14   \\
ABMIL            & 7.86M  & 12.84  \\
HE2RNA           & 15.97M & 22.40  \\
AbReg            & 7.66M  & 149.86 \\
tRNAformer       & 5.22M  & \textbf{1.86}   \\
ILRA             & 6.77M  & 33.42  \\
S4MIL            & 1.91M  & \underline{1.96}   \\
MambaMIL         & 2.00M  & 3.12   \\
SRMambaMIL       & 2.02M  & 3.76   \\
MOSBY            & 8.25M  & 161.12 \\
SEQUOIA VIS      & 20.68M & 2.76   \\
2DMamba          & 3.99M  & 516.02 \\
CPNN             & \underline{1.51M}  & 28.76  \\
\midrule
MoSPR (ours)     & \textbf{0.96M} & 2.02 \\
\bottomrule
\end{tabular}
\end{table}

\begin{table}[!h]
\caption{
Low-rank output controls for ABMIL and tRNAformer.
Values report mean gene SCC over four cross-validation folds.
\emph{Low-Rank} fits each baseline in its own validation-selected
$q$-dimensional PCA target space, whereas \emph{Post-Hoc} projects the
full-rank predictions onto the same subspace ($U_q^\top U_q$) without
retraining. The rank $q$ is selected independently for each baseline and
fold; MoSPR is shown for reference.
}
\label{tab:supp-lowrank-control}
\centering
\small
\setlength{\tabcolsep}{6pt}
\renewcommand{\arraystretch}{1.0}

\begin{tabular}{llccc}
\toprule
Cohort & Model & Full-Rank& Low-Rank& Post-Hoc\\
\midrule
\multirow{3}{*}{TCGA-BRCA}
 & ABMIL & 0.363 & 0.354 & 0.380 \\
 & tRNAformer & 0.332 & 0.327 & 0.338 \\
 & MoSPR (ours) & \multicolumn{3}{c}{\textbf{0.411}} \\
\midrule
\multirow{3}{*}{TCGA-KIRC}
 & ABMIL & 0.299 & 0.322 & 0.314 \\
 & tRNAformer & 0.319 & 0.308 & 0.324 \\
 & MoSPR (ours) & \multicolumn{3}{c}{\textbf{0.348}} \\
\midrule
\multirow{3}{*}{TCGA-LUAD}
 & ABMIL & 0.324 & 0.326 & 0.345 \\
 & tRNAformer & 0.336 & 0.324 & 0.344 \\
 & MoSPR (ours) & \multicolumn{3}{c}{\textbf{0.376}} \\
\bottomrule
\end{tabular}
\end{table}

\begin{table}[!h]
\caption{
Macrostate partition stability under patient-level bootstrap resampling.
Microstates were fixed, and 300 patient-level bootstrap resamples were generated
for each cancer cohort. For each resample, the global microstate adjacency matrix
was reconstructed and re-clustered into $K=8$ macrostates. Partition stability
was quantified using the Adjusted Rand Index (ARI; \citealp{hubert1985comparing}) between the bootstrap-derived
macrostate partition and the reference partition obtained from the full cohort.
}
\label{tab:supp-macrostate-ari}
\centering
\small
\setlength{\tabcolsep}{6pt}
\renewcommand{\arraystretch}{1.0}

\begin{tabular}{lccc}
\toprule
Cohort & Patients& Mean ARI& Median ARI\\
\midrule
TCGA-BRCA & 1{,}037 & 0.769 & 0.801  \\
TCGA-KIRC & 342     & 0.653 & 0.646  \\
TCGA-LUAD & 482     & 0.786 & 0.798 \\
\bottomrule
\end{tabular}
\end{table}

\clearpage

\section{Supplementary Figures}

\setcounter{figure}{0}
\renewcommand{\thefigure}{S\arabic{figure}}
\renewcommand{\theHfigure}{S\arabic{figure}}

\begin{figure}[!h]
    \centering
    \includegraphics[width=0.8\linewidth]{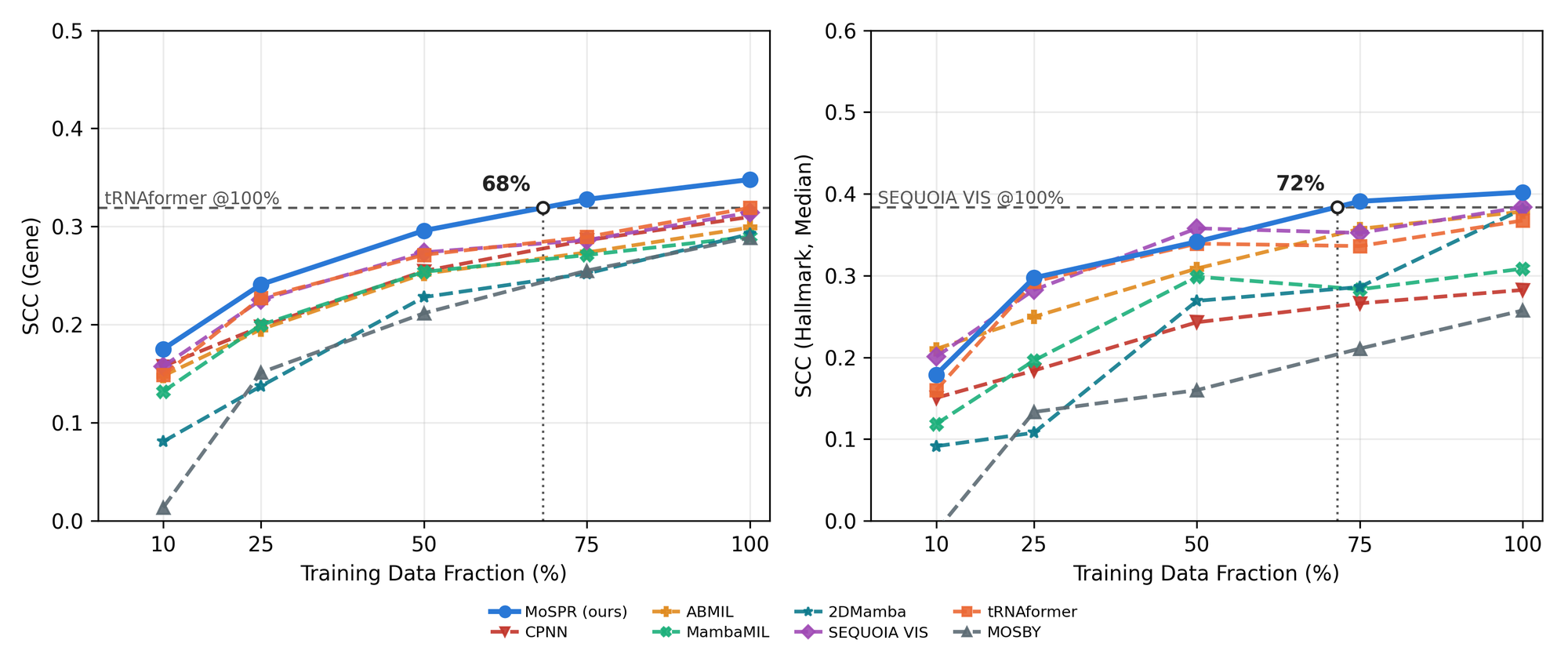}
    \caption{
    Training-set scaling on TCGA-KIRC.
    Gene-expression prediction performance (left) and Hallmark pathway
    reconstruction performance (right) are shown across increasing fractions
    of the training data.
    }
    \label{fig:supp-data-efficiency-kirc}
\end{figure}

\begin{figure}[!h]
    \centering
    \includegraphics[width=0.8\linewidth]{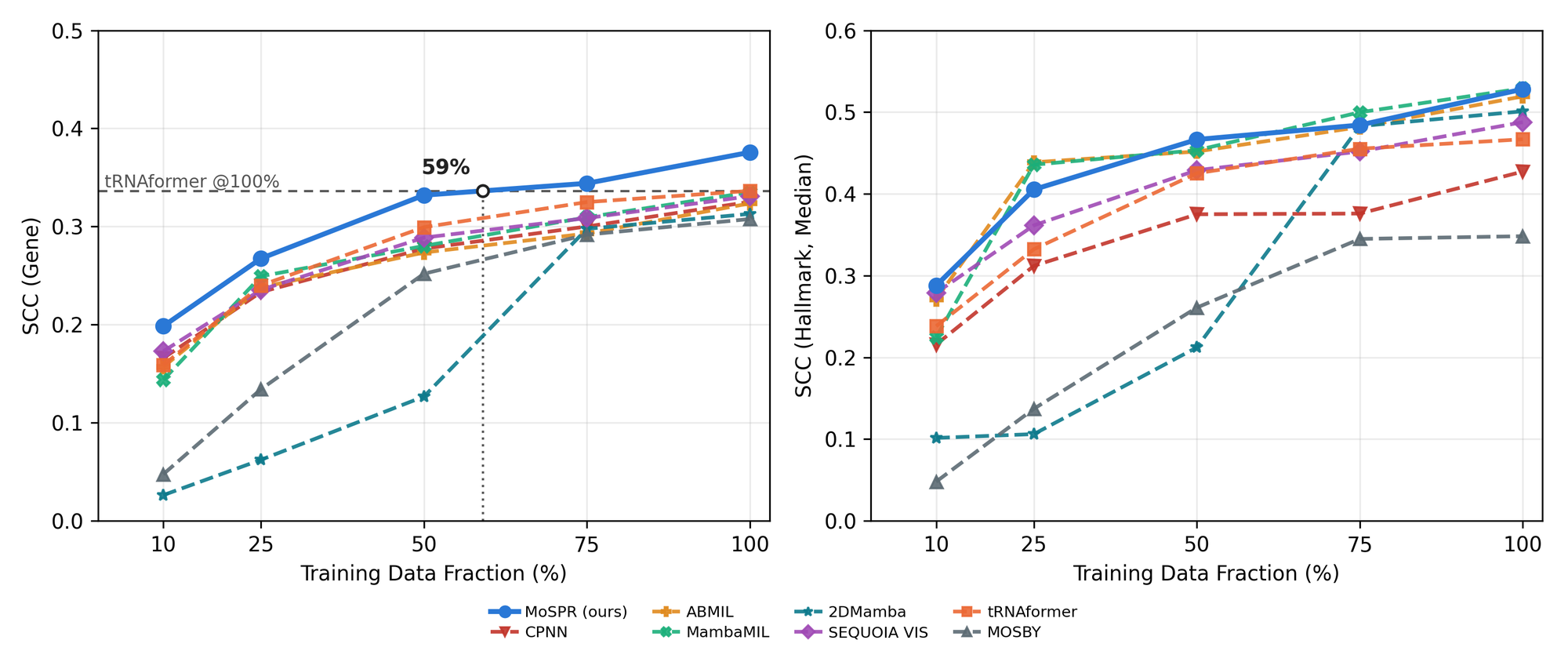}
    \caption{
    Training-set scaling on TCGA-LUAD.
    Gene-expression prediction performance (left) and Hallmark pathway
    reconstruction performance (right) are shown across increasing fractions
    of the training data.
    }
    \label{fig:supp-data-efficiency-luad}
\end{figure}

\end{document}